\documentclass[runningheads]{llncs}
\usepackage{multirow}
\usepackage{graphicx}
\usepackage{amsmath}

\newcommand{\bc}{{\bf c}}
\newcommand{\bd}{{\bf d}}
\newcommand{\bo}{{\bf o}}
\newcommand{\br}{{\bf r}}
\newcommand{\bx}{{\bf x}}

\newcommand{\fcaption}[2]{\caption{{\bf #1}. {#2}}}

\newcommand{\fig}[1]{Fig.~\ref{fig:#1}}

\newcommand{\tab}[1]{Table~\ref{tab:#1}}

\newcommand{\sect}[1]{Section~\ref{sec:#1}}

\newcommand{\eq}[1]{Eq.~\eqref{eq:#1}}

\usepackage{xcolor}

\newcommand{\changed}[1]{#1}

\usepackage{hyperref}
\usepackage[pscoord]{eso-pic}
\newcommand{\placetextbox}[3]{
  \setbox0=\hbox{#3}
  \AddToShipoutPictureFG*{
    \put(\LenToUnit{#1\paperwidth},\LenToUnit{#2\paperheight}){\vtop{{\null}\makebox[0pt][c]{\begin{tabular}{l}#3\end{tabular}}}}%
  }%
}%

\begin{document}
\placetextbox{0.5}{.96}{\small
\textit{This version of the contribution has been accepted for publication, after peer review but is not} \\
\small \textit{the Version of Record and does not reflect post-acceptance improvements, or any corrections.} \\
\small \textit{The Version of Record is available online at: \href{http://doi.org/10.1007/978-3-031-43999-5_48}{http://doi.org/10.1007/978-3-031-43999-5\_48}.} \\
\small \textit{Use of this Accepted Version is subject to the publisher's Accepted Manuscript terms of use} \\
\small \textit{\href{https://www.springernature.com/gp/open-research/policies/accepted-manuscript-terms}{https://www.springernature.com/gp/open-research/policies/accepted-manuscript-terms}}}
\title{LightNeuS: Neural Surface Reconstruction in Endoscopy using Illumination Decline}

\titlerunning{LightNeus: Neural Surface Reconstruction in Endoscopy}
\iftrue
\author{V\'ictor M. Batlle\inst{1}\and Jos\'e M. M. Montiel\inst{1}\and Pascal Fua\inst{2}\and Juan D. Tard\'os\inst{1}}
\authorrunning{V. M. Batlle et al.}
\institute{Inst. Investigaci\'on en Ingenier\'ia de
Arag\'on, I3A, Universidad de Zaragoza, Spain\\
\email{\{vmbatlle,josemari,tardos\}@unizar.es}
\and
CVLab, \'Ecole Polytechnique F\'ed\'erale de Lausanne, Switzerland\\
\email{pascal.fua@epfl.ch}}
\else
\author{Anonymous}
\authorrunning{Anonymous}
\fi
\maketitle              

\begin{abstract}
    
We propose a new approach to 3D reconstruction from sequences of images acquired by monocular endoscopes. It is based on two key insights. First, endoluminal cavities are watertight,  a property  naturally enforced by modeling them in terms of a signed distance function. Second, the scene illumination is variable. It comes from the endoscope's light sources and decays with the inverse of the squared distance to the surface. To exploit these insights, we build on NeuS~\cite{wang2021neus}, a neural implicit surface reconstruction technique with an outstanding capability to learn appearance and a SDF surface model from multiple views, but currently limited to scenes with static illumination. To remove this limitation and exploit the relation between pixel brightness and depth, we modify the NeuS architecture to explicitly account for it and  introduce a calibrated photometric model of the endoscope's camera and light source.  

Our method is the first one to produce watertight reconstructions of whole colon sections.
We demonstrate excellent accuracy on phantom imagery. 
Remarkably, the watertight prior combined with illumination decline, allows to complete the reconstruction of unseen portions of the surface with acceptable accuracy, paving the way to automatic quality assessment of cancer screening explorations, measuring the global percentage of observed mucosa.

\end{abstract}
 
\keywords{Reconstruction \and Photometric multi-view \and Endoscopy}

\section{Introduction}
\label{sec:intro}

Colorectal cancer (CRC) is the third most commonly diagnosed cancer and is the second most common cause of cancer death \cite{sung2021global}. Early detection is crucial for a good prognosis. Despite the existence of other techniques, such as virtual colonoscopy (VC), optical colonoscopy (OC) remains the gold standard for colonoscopy screening and the removal of precursor lesions. Unfortunately, we do not yet have the ability to reconstruct densely the 3D shape of large sections of the colon. This would usher exciting new developments, such as post-intervention diagnosis, measuring polyps and stenosis, and automatically evaluating exploration thoroughness in terms of the surface percentage that has been observed.

This is the problem we address here. It has been shown that the colon 3D shape can be estimated from single images acquired during human colonoscopies~\cite{batlle2022photometric}. However, to model large sections of it while increasing the reconstruction accuracy, multiple images must be used. As most endoscopes contain a single camera, the natural way to do this is to use video sequences acquired by these cameras in the manner of structure-from-motion algorithms. An important first step in that direction is to register the images from the sequences. This can now be done reliably using either batch~\cite{schoenberger2016sfm} or SLAM techniques~\cite{gomez2021sd}.
Unfortunately, this solves only half the problem because these techniques provide very sparse reconstructions and going from there to dense ones remains an open problem. And  occlusions, specularities, varying albedos, and specificities of endoscopic lighting make it a challenging one.

To overcome these difficulties, we rely on two properties of endoscopic images: 
\begin{itemize}
    \item Endoluminal cavities such as the gastrointestinal tract, and in particular the human colon, are watertight surfaces. To account for this, we represent its surface in terms of a signed distance function (SDF), which by its very nature presents continuous watertight surfaces.
    
   \item In endoscopy the light source is co-located with the camera. It  illuminates a dark scene and is always close to the surface.  As a result, the irradiance decreases rapidly with distance $t$ from camera to surface; more specifically  it is a function of  $1/t^2$. In other words, there is a strong correlation between light and depth, which remains unexploited to date.
           
\end{itemize}
To take advantage of these specificities, we build on the success of Neural implicit Surfaces (NeuS)~\cite{wang2021neus} that have been shown to be highly effective at deriving surface 3D models from sets of registered images. As the Neural Radiance Fields (NeRFs)~\cite{mildenhall2021nerf} that inspired them, they were designed to operate on regular images taken around a scene, sampling fairly regularly the set of possible viewing directions. Furthermore, the lighting is assumed to be static and distant so that the brightness of a pixel and its distance to the camera are unrelated. Unfortunately, none of these conditions hold in endoscopies. The camera is inside a cavity (in the colon, a roughly cylindrical tunnel) that limits viewing directions. The light source is co-located with the camera and close to the surface, which results in a strong correlation between pixel brightness and distance to the camera. In this paper, we show that, far from being a handicap, this correlation is a key information for neural network self-supervision. 

\begin{figure}[!tb]
    \begin{center}
    \includegraphics[width=0.9\textwidth]{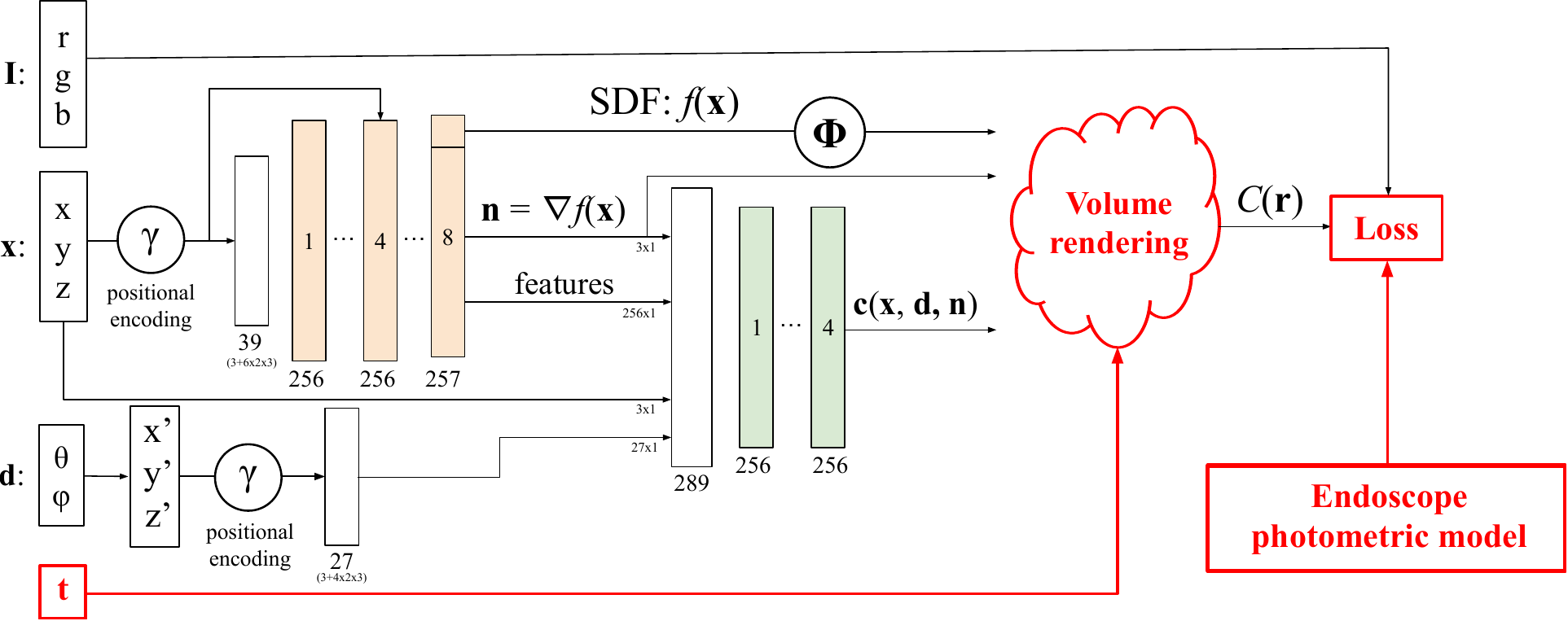}
    \end{center}
\fcaption{From NeuS to LightNeuS} {The original NeuS architecture is depicted by the black arrows. In LightNeuS, \changed{when training the network with a sampled point, we provide the sampling distance $t$ to the renderer}, that takes into account illumination decline. We also incorporate a calibrated photometric endoscope model that is used to correctly compute the photometric loss. The changes are shown in red. }
    \label{fig:neus}
\end{figure}

\changed{NeuS training selects a pixel from an image and samples points along its projecting ray. However, the network is agnostic to the sampling distance. In LightNeuS, we explicitly feed to the renderer the distance of each one of these sampled points to the light source, as shown in Fig.~\ref{fig:neus}. Hence, the renderer can exploit the inverse-square illumination decline.}  We also introduce and calibrate a photometric model for the endoscope light and camera, so that the inverse square law discussed above actually holds. \changed{Together, these two changes make the minimization problem better posed and the automatic depth estimation more reliable.}

Our results show that exploiting the illumination is key to unlocking implicit neural surface reconstruction in endoscopy. \changed{It delivers accuracies in the range of 3 mm, whereas an unmodified NeuS is either 5 times less accurate or even fails to reconstruct any surface at all. Earlier methods~\cite{batlle2022photometric} have reported similar accuracies but only on very few synthetic images and on short sections of the colon. By contrast, we can handle much longer ones and provide a broad evaluation in a real dataset (C3VD) over multiple sequences.  This makes us the first to show accurate results of extended 3D watertight surfaces from monocular endoscopy images.}

\section{Related Works}
\label{sec:related}

\textbf{3D Reconstruction from Endoscopic Images.}
It can help with the effective localization of lesions, such as polyps and adenomas, by providing a complete representation of the observed surface. Unfortunately, many state-of the-art SLAM techniques based on feature matching \cite{campos2021orb} or direct methods \cite{engel2014lsd,engel2017direct} are impractical for dense endoscopic reconstruction due to the lack of texture and the inconsistent lighting that moves along with the camera. Nevertheless, sparse reconstructions by classical Structure-from-Motion (SfM) algorithms can be good starting points for refinement and densification based on Shape-from-Shading (SfS) \cite{tokgozoglu2012color,zhao2016endoscopogram}. However, classical multi-view and SfS methods require strong suboptimal priors on colon surface shape and reflectance.

In monocular dense reconstructions, it is common practice to encode shape priors in terms of smooth rigid surfaces \cite{Newcombe2011,schonberger2016pixelwise,Mahmoud2019}. Recently, \cite{sengupta2021colonoscopic} proposes a tubular topology prior for NRSfM aimed to process endoluminal cavities where these tubular shapes are prevalent. In contrast, for the same environments, we propose the watertight prior coded by implicit SDF representations.

Recent methods for dense reconstruction rely on neural networks to predict per-pixel depth in the 2D space of each image and fuse the depth maps by using multi-view stereo (MVS) \cite{bae2020deep} or a SLAM pipeline \cite{ma2019real,ma2021rnnslam}. However, holes in the reconstruction appear due to failures in triangulation and inaccurate depth estimation or in areas not observed in any image. Wang et al. \cite{wang2022neural} show the potential of neural rendering in reconstruction from medical images, although they use a binocular static camera with fixed light source, which is not feasible in endoluminal endoscopy. \changed{Unfortunately, most of the previous 3D methods do not provide code \cite{Mahmoud2019,sengupta2021colonoscopic}, are not evaluated in biomedical settings \cite{Newcombe2011,schonberger2016pixelwise}, or do not report reconstruction accuracy \cite{ma2019real,ma2021rnnslam}.}

\bigbreak

\textbf{Neural Radiance Fields (NeRFs)}
were first proposed to reconstruct novel views of non-Lambertian objects~\cite{mildenhall2021nerf}. This method provides an \emph{implicit neural representation} of a scene in terms of local densities and associated colors. In effect, the scene representation is stored in the weights of a neural network, usually a multilayer perceptron (MLP), that learns its shape and reflectance for any coordinate and viewing direction. NeRFs use volume rendering \cite{kajiya1984ray}, based on ray-tracing from multiple camera positions. The volume density $\sigma(\bx)$ can be interpreted as the differential probability of a ray terminating at an infinitesimal particle at location $\bx$. The expected color $C(\br)$ of the pixel with camera ray $\br(t)= \bo + t {\bf d}$ is the integration of the radiance emitted by the field at every traveled distance $t$ from near to far bounds $t_n$ and $t_f$, such that
\begin{small}
\begin{equation}
C({\bf r}) =\int_{t_n}^{t_f} T(t) \; \sigma(\br(t)) \; {\bf c}({\bf r}(t),\bd)  \, \mathrm{d} t \;\;\; \mbox{ where } T(t) = \exp \left( - \int_{t_n}^t \sigma (\br(s)) \, \mathrm{d} s \right) \; 
\label{eq:nerf}
\end{equation}
\end{small}

\noindent where $\bc$ stands for the color. The function $T$ denotes the accumulated transmittance along the ray from $t_n$ to $t$, that is the probability that the ray travels from $t_n$  to $t$ without hitting any other particle. The authors propose two MLPs to estimate the volume density function $\sigma: \bx \rightarrow [0, 1]$ and the directional emitted color function $\mathbf{c}: (\bx, \bd) \rightarrow [0, 1]^3$, so the density of a point does not depend on the viewing direction $\bd$, but the color does. This allows them to model non-Lambertian reflectance. In addition, they propose a positional encoding for location $\bx$ and direction $\bd$, which allows high-frequency details in the reconstruction.

\bigbreak

\textbf{Neural Implicit Surfaces (NeuS)}
were  introduced in~\cite{wang2021neus}  to improve the quality of NeRF representation modelling watertight surfaces. For that, the volume density  $\sigma$ is computed so as to be maximal at the zero-crossings of a signed distance function (SDF) $f$:
\begin{equation}
\sigma(\br(t)) = \max \left(\frac{\Phi_s'(f(\br(t)))}{\Phi_s(f(\br(t)))},0\right) \;\; \mbox{ where } \Phi_s(x) = \frac{1}{1 + e^{-sx}}
\label{eq:neus}
\end{equation}

The SDF formulation makes it possible to estimate the surface normal as $\mathbf{n} = \nabla f(\bx)$. The reflectance of a material is usually determined as a function of the incoming and outgoing light directions with respect to the surface normal. Therefore, the normal is added as an input to the MLP that estimates color $\mathbf{c} : (\mathbf{x}, \mathbf{d}, \mathbf{n})$, as shown in \fig{neus}.
 \section{LightNeuS}
\label{sec:method}

In this section, we present the key contributions that make {\it LightNeuS} a neural implicit reconstruction method suitable for endoscopy in endoluminal cavities. In this context, the light source is located next to the camera and moves with it. Furthermore, it is close to the surfaces to be modeled. As a result, for any surface point $ \mathbf{x} = \mathbf{o} + t\mathbf{d} $ , the irradiance decreases with the square of the distance to the camera $t$. Hence, we can write the color of the corresponding  pixel as \changed{\cite{batlle2022photometric}}:
\begin{equation}
\label{eq:render}
    \mathcal{I}(\mathbf{x})
    =
    {\left(
        \frac{L_e}{t^2} \;
        \text{BRDF}(\mathbf{x}, \mathbf{d}) \;
        \cos\left(\theta \right)\;
        g
    \right)}^{1/\gamma} 
\end{equation}
where $L_e$ is the radiance emitted by the light source to the surface point, \changed{that was modeled and calibrated in the EndoMapper dataset \cite{azagra2022} according to the SLS model from \cite{modrzejewski2020light}.} The bidirectional reflectance distribution function (BRDF) determines how much light is reflected to the camera, and the cosine term $\cos\left(\theta\right) = -\mathbf{d} \cdot \mathbf{n}$ weights the incoming radiance with respect to the surface normal $\mathbf{n}$. Equation~(\ref{eq:render})  also takes into account the camera gain $g$ and gamma correction $\gamma$.

\subsection{Using Illumination Decline as a Depth Cue}

The NeuS formulation of \sect{related} assumes distant and fixed lighting. However, in endoscopy inverse-square light decline is significant, as quantified in \eq{render}.
Accounting for this is done by modifying the original NeuS formulation as follows. \fig{neus} depicts the original NeuS network in black. It uses a SDF network---shown in orange---to estimate a view-independent geometry and only the final RGB color depends on the viewing direction $\mathbf{d}$. It is estimated by the network shown in green. Thus, this second network $\mathbf{c}(\mathbf{x}, \mathbf{d}, \mathbf{n})$ may learn to model non-Lambertian BRDF$(\mathbf{x}, \mathbf{d})$, including  specular highlights, and the cosine term of \eq{render}. However, if the distance $t$ from the light to the point $\bx$ is not provided to the color network, the ${1}/{t^2}$ dependency cannot be learned, and surface reconstruction will fail. Our key insight is to explicitly supply this distance as input to the volume rendering algorithm, as shown in red in \fig{neus} and reformulate \eq{nerf} as
\begin{equation}
\label{eq:ours}
    C({\bf r}) = \int_{t_n}^{t_f} T(t)\; \sigma(\br(t)) \; \frac{{\bc}({\br}(t),\bd,\mathbf{n})}{t^2}  \, \mathrm{d} t 
\end{equation}
This conceptually simple change, using illumination decline while training, unlocks all the power of neural surface reconstruction in endoscopy.  

\subsection{Endoscope Photometric Model}

Apart from illumination decline, there are several significant differences between the images captured by endoscopes and those conventionally used to train NeRFs and NeuS: fish-eye lenses, strong vignetting, uneven scene illumination,  and post-processing.  

Endoscopes use fisheye lenses to cover a wide field of view, usually close to 170 degrees. These lenses produce strong deformations, making it unwise to use the standard pinhole camera model. Instead, specific models \cite{scaramuzza2006toolbox,kannala2006generic} must be used. Hence, we also modified the original NeuS implementation to support these models.

The light sources of endoscopes behave like spotlights. In other words, they do not emit with the same intensity in all directions, so $L_e$ in \eq{render} is not constant for all image pixels. This effect is similar to the vignetting effect caused by conventional lenses, that is aggravated in fisheye lenses. Fortunately, they can be accurately calibrated~\cite{azagra2022,modrzejewski2020light} and compensated for.

The post-processing software of medical endoscopes is designed to always display well-exposed images, so that physicians can see details correctly. An adaptive gain factor $g$ is applied by the endoscope's internal logic and gamma correction is also used to adapt to non-linear human vision, achieving better contrast perception in mid tones and  dark areas. Endoscope manufacturers know the post-processing logic of their devices, but this information is proprietary and not available to users. Again, gamma correction can be calibrated assuming it is constant \cite{batlle2022photometric}, and the gain change between successive images can be estimated, for example, by sparse feature matching.

All these factors must be taken into account during network training.
Thus, our photometric loss is computed using a normalized image:
\begin{equation}
I' = \left(\frac{I^\gamma}{L_e g}\right)^{1/\gamma} 
\end{equation}
 \section{Experiments}

\changed{We validate our method on the C3VD dataset \cite{bobrow2022colonoscopy}, which covers all different sections of the colon anatomy in 22 video sequences.} This dataset contains sequences recorded with a medical video colonoscope, Olympus Evis Exera III CF-HQ190L. The images were recorded inside a \emph{phantom}, a model of a human colon made of silicone. The intrinsic camera parameters are provided. The camera extrinsics for each frame are estimated by 2D-3D registration against the known 3D model. In an operational setting, we could use a structure-from-motion approach such as COLMAP~\cite{schoenberger2016sfm} or a SLAM technique such as~\cite{gomez2021sd}, which have been shown to work well in endoscopic settings.
The gain values were easily estimated from the dataset itself. For vignetting, we use the calibration obtained from a colonoscope of the same brand and series from the EndoMapper dataset \cite{azagra2022}.

\changed{During training, we follow the NeuS paper approach of using a few informative frames per scene, as separated as possible, by sampling each video uniformly.} For each sequence, we train both the vanilla NeuS and our LighNeuS using 20 frames each time. They are extracted uniformly over the duration of the video. We use the same batch size and number of iterations as in the original NeuS paper, 512 and 300k respectively. Once the network is trained, we can extract triangulated meshes from the reconstruction.  Since the C3VD dataset comprises a ground-truth triangle mesh, we compute point-to-triangle distances from all the vertices in the reconstruction to the closest ground-truth triangle.

\begin{table}[!t]
\caption{\changed{{\bf Reconstruction error [mm] on the C3VD dataset.} {\bf \underline{Sur}veyed:} points seen at least once. {\bf \underline{Ext}ended:} points within 20~mm of a visible point. Anatomical regions: \underline{C}ecum, \underline{D}escending, \underline{S}igmoid and \underline{T}ransverse. For NeuS, we provide two sets of numbers because the optimization failed on the other sections. In \textit{italics} we mark the sequences where the camera moves less than 1 cm yielding higher errors.}}
\label{tab:metrics}
\begin{tabular}{ccccccccccccccc}
\cline{3-15}
 & \multicolumn{1}{c|}{} & \multicolumn{2}{c|}{\textbf{NeuS}} & \multicolumn{11}{c}{\textbf{LightNeuS (ours)}} \\ \hline
\multicolumn{2}{c|}{\textbf{Sequence}} & \textbf{C1a} & \multicolumn{1}{c|}{\textbf{C4b}} & \textbf{C1a} & \textbf{C1b} & \textbf{C2a} & \textbf{C2b} & \textbf{C2c} & \textbf{C3a} & \textbf{C4a} & \textbf{C4b} & \textbf{D4a} & \textbf{S1a} & \textbf{S2a} \\ \hline
\parbox[t]{4mm}{\multirow{3}{*}{\rotatebox[origin=c]{90}{\textbf{Sur.}}}} & \multicolumn{1}{c|}{MedAE} & 4.53 & \multicolumn{1}{c|}{10.6} & 0.95 & 4.85 & 1.40 & 3.26 & 2.57 & 1.12 & 1.90 & 1.41 & 2.66 & 4.23 & 1.19 \\
 & \multicolumn{1}{c|}{MAE} & 5.07 & \multicolumn{1}{c|}{10.6} & 1.48 & 5.11 & 1.54 & 3.65 & 3.00 & 2.54 & 2.14 & 1.63 & 3.26 & 4.33 & 1.89 \\
 & \multicolumn{1}{c|}{RMSE} & 6.40 & \multicolumn{1}{c|}{11.6} & 2.01 & 5.63 & 1.87 & 4.39 & 3.74 & 5.49 & 2.92 & 2.10 & 4.08 & 4.96 & 2.78 \\ \hline
\parbox[t]{4mm}{\multirow{3}{*}{\rotatebox[origin=c]{90}{\textbf{Ext.}}}} & \multicolumn{1}{c|}{MedAE} & 4.68 & \multicolumn{1}{c|}{5.35} & 0.83 & 4.89 & 1.41 & 3.32 & 2.54 & 1.27 & 1.91 & 1.45 & 4.50 & 4.01 & 1.40 \\
 & \multicolumn{1}{c|}{MAE} & 6.24 & \multicolumn{1}{c|}{6.74} & 1.26 & 5.10 & 1.56 & 3.70 & 3.01 & 3.83 & 2.18 & 1.72 & 6.61 & 4.19 & 2.36 \\
 & \multicolumn{1}{c|}{RMSE} & 8.77 & \multicolumn{1}{c|}{8.56} & 1.72 & 5.60 & 1.90 & 4.42 & 3.77 & 7.96 & 2.95 & 2.20 & 9.32 & 4.87 & 3.96 \\ \hline
 &  &  &  &  &  &  &  &  &  &  &  &  &  &  \\ \cline{3-15} 
\multicolumn{2}{c}{\textbf{}} & \multicolumn{13}{c|}{\textbf{LightNeuS (ours)}} \\ \cline{3-15} 
\multicolumn{2}{c}{\textbf{}} & \textbf{S3a} & \textbf{S3b} & \textbf{T1a} & \textbf{T1b} & \textbf{T2a} & \textbf{T2b} & \multicolumn{1}{c|}{\textbf{T4a}} & \multicolumn{1}{c|}{\textbf{Mean}} & \textit{\textbf{T2c}} & \textit{\textbf{T3a}} & \textit{\textbf{T3b}} & \multicolumn{1}{c|}{\textit{\textbf{T4b}}} & \multicolumn{1}{c|}{\textit{\textbf{Mean}}} \\ \cline{3-15} 
 &  & 2.57 & 3.63 & 3.43 & 2.33 & 2.24 & 2.16 & \multicolumn{1}{c|}{1.15} & \multicolumn{1}{c|}{\textbf{2.39}} & \textit{5.07} & \textit{6.39} & \textit{11.0} & \multicolumn{1}{c|}{\textit{1.75}} & \multicolumn{1}{c|}{\textit{\textbf{6.04}}} \\
 &  & 2.68 & 4.16 & 3.47 & 2.72 & 2.28 & 2.30 & \multicolumn{1}{c|}{2.31} & \multicolumn{1}{c|}{\textbf{2.80}} & \textit{5.45} & \textit{8.65} & \textit{12.1} & \multicolumn{1}{c|}{\textit{6.70}} & \multicolumn{1}{c|}{\textit{\textbf{8.23}}} \\
 &  & 3.18 & 4.81 & 4.07 & 3.34 & 2.58 & 2.70 & \multicolumn{1}{c|}{3.79} & \multicolumn{1}{c|}{\textbf{3.58}} & \textit{6.48} & \textit{10.7} & \textit{14.4} & \multicolumn{1}{c|}{\textit{11.3}} & \multicolumn{1}{c|}{\textit{\textbf{10.7}}} \\ \cline{3-15} 
 &  & 2.87 & 3.54 & 3.38 & 2.69 & 2.19 & 2.12 & \multicolumn{1}{c|}{1.29} & \multicolumn{1}{c|}{\textbf{2.53}} & \textit{4.44} & \textit{6.54} & \textit{13.6} & \multicolumn{1}{c|}{\textit{8.00}} & \multicolumn{1}{c|}{\textit{\textbf{8.16}}} \\
 &  & 3.27 & 4.64 & 3.31 & 3.21 & 2.22 & 2.28 & \multicolumn{1}{c|}{2.22} & \multicolumn{1}{c|}{\textbf{3.15}} & \textit{5.36} & \textit{8.10} & \textit{14.1} & \multicolumn{1}{c|}{\textit{10.4}} & \multicolumn{1}{c|}{\textit{\textbf{9.47}}} \\
 &  & 4.04 & 6.10 & 3.86 & 3.96 & 2.55 & 2.69 & \multicolumn{1}{c|}{3.32} & \multicolumn{1}{c|}{\textbf{4.18}} & \textit{6.78} & \textit{9.94} & \textit{15.9} & \multicolumn{1}{c|}{\textit{13.9}} & \multicolumn{1}{c|}{\textit{\textbf{11.6}}} \\ \cline{3-15} 
\end{tabular}
\end{table} 
\changed{In the first rows} of \tab{metrics}, we report median (MedAE), mean (MAE), and root mean square (RMSE) values of these distances for all vertices seen in at least one image. \changed{Columns show the result for 22 sequences. We note 18 sequences where the camera moved at least 1~cm, and the reconstruction yielded a mean error of 2.80~mm. The other four smaller trajectories ($<$1cm) lack parallax and the mean error is higher (8.23mm).}

This is in the range of reported accuracy in the literature for monocular dense non-watertight depth estimation, 1.1\,mm in \cite{Mahmoud2019} for high parallax geometry in laparoscopy, which is a much more favorable geometry than the one we have here, or  0.85\,mm for the significantly smaller-size cavities of endoscopic endonasal surgery (ESS)~\cite{Liu2022}. 

\changed{In contrast, vanilla NeuS assumes constant illumination. The strong light changes typical of endoscopy fatally mislead the method. We only report numerical results of NeuS in two sequences because in all the rest, the SDF diverges and ends up blown out of the rendering volume, giving no result at all.}

\begin{figure}[!t]
    \centering
    \begin{tabular}{cccc}
        & 3D Reconstruction & Error & \\
        \rotatebox[origin=l]{90}{\qquad \quad NeuS} &
            \includegraphics[width=0.2\textwidth]{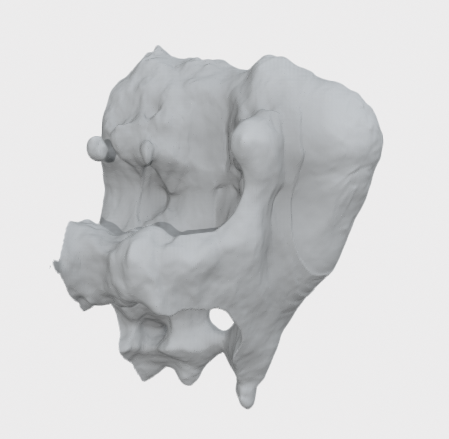} & \includegraphics[width=0.2\textwidth]{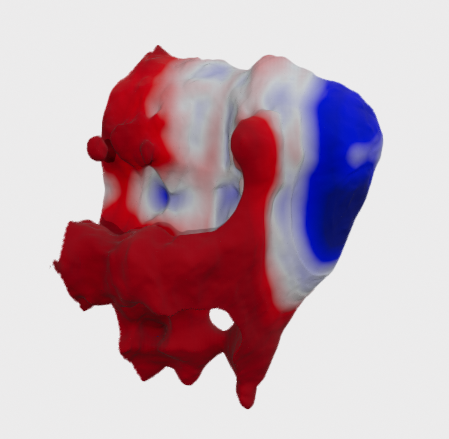} &
            \multirow{2}{*}[5.5em]{\begin{tabular}{@{}c@{}}Ground truth (GT) \\
                \includegraphics[width=0.2\textwidth]{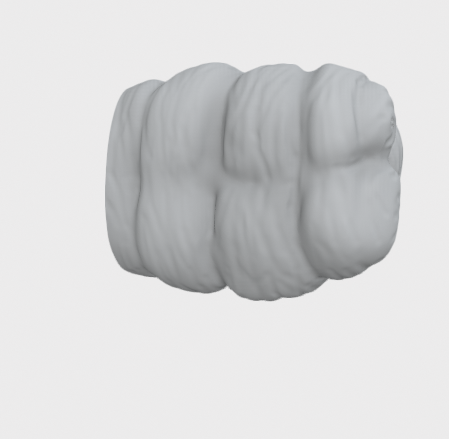 } \\
                \includegraphics[width=0.2\textwidth]{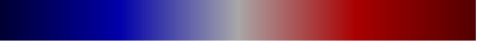} \\
                -1~cm \, \quad 0 \, \quad +1~cm
                \end{tabular}}  \\
        \rotatebox[origin=l]{90}{\quad \quad LigthNeuS} &
            \includegraphics[width=0.2\textwidth]{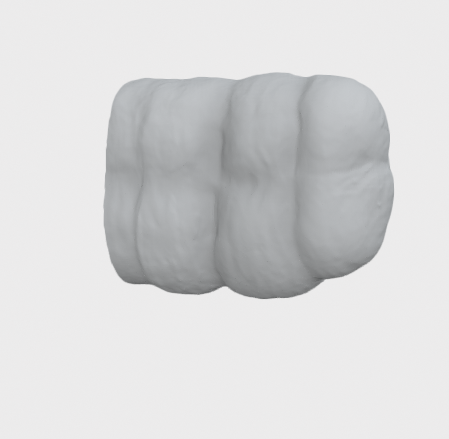} & \includegraphics[width=0.2\textwidth]{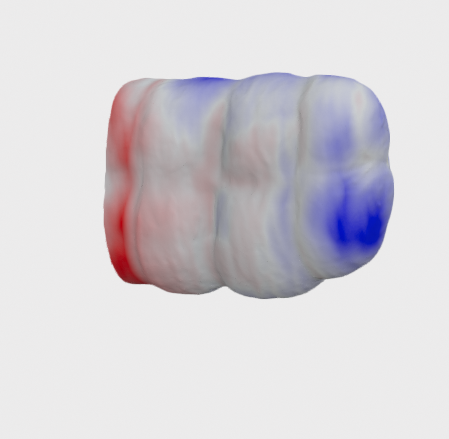} & 
    \end{tabular}
\fcaption{Benefits of illumination decline}{Result on the \emph{``Cecum 1 a''} sequence. {\bf Top:} The NeuS reconstruction exhibits multiple artifacts that make it unusable. {\bf Bottom:} Our reconstruction is much closer to the ground truth shape. The error is shown in blue if the reconstruction is inside the surface, and in red otherwise. A fully saturated red or blue denotes an error of more than 1cm and grey denotes no error at all.}
    \label{fig:with_without}
\end{figure} 

\begin{figure}
    \centering
    \begin{tabular}{ccc}
        (a) First frame & (c) Ground Truth & (e) 3D Reconstruction \\
        \includegraphics[width=0.2\textwidth]{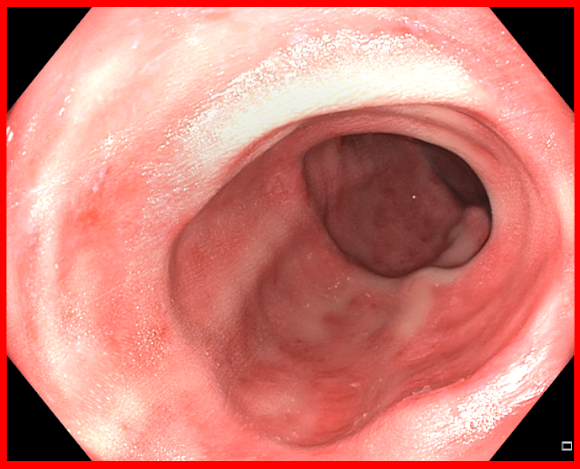} & 
        \includegraphics[width=0.2\textwidth]{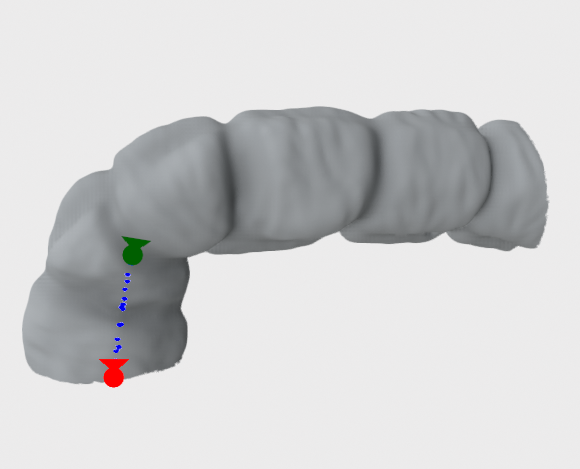} &
        \includegraphics[width=0.2\textwidth]{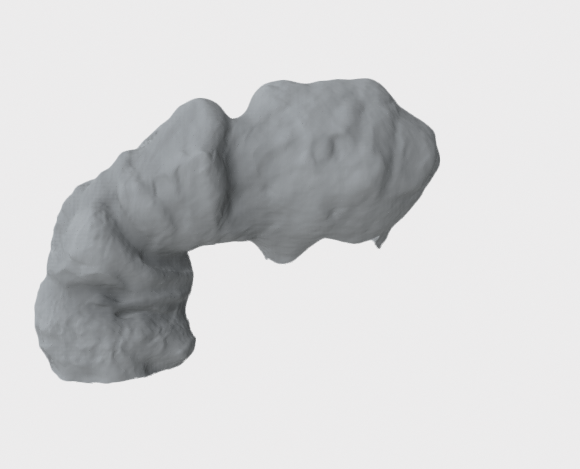} \\
        (b) Deepest frame & (d) Surveyed (GT) & (f) Error \\
        \includegraphics[width=0.2\textwidth]{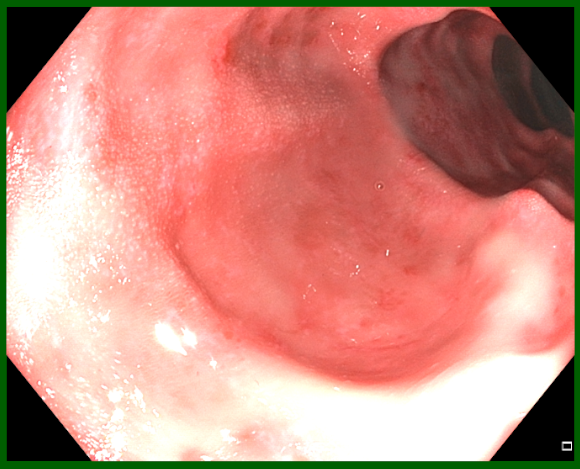} &
        \includegraphics[width=0.2\textwidth]{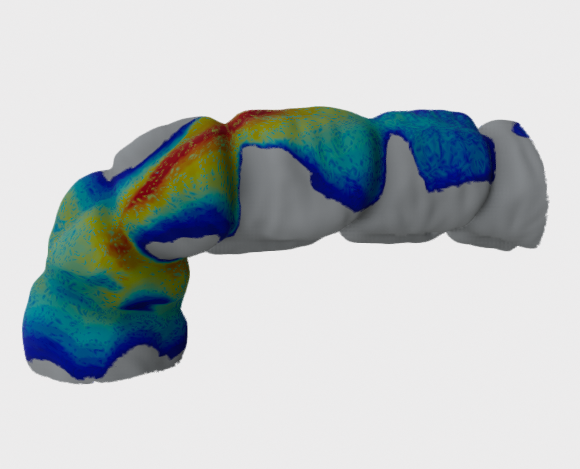} &
        \includegraphics[width=0.2\textwidth]{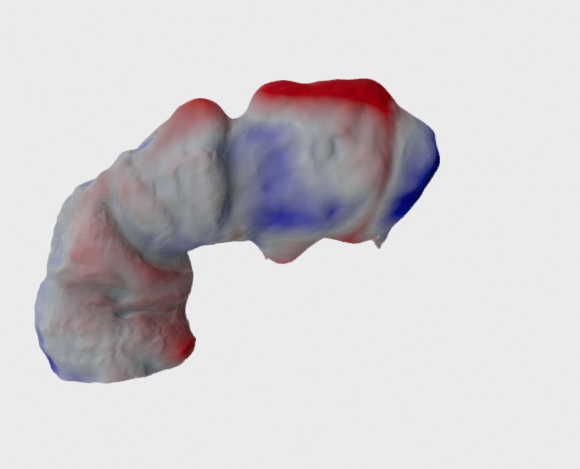} \\
        & \includegraphics[width=0.2\textwidth]{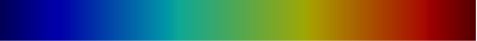} & 
        \includegraphics[width=0.2\textwidth]{fig/colorbar_seismic.jpg} \\
        & 1 \quad \qquad \;\; \qquad \quad 20 & -1~cm \, \quad 0 \, \quad +1~cm
        
    \end{tabular}\fcaption{Reconstructing partially observed regions}{Results on \emph{``Transcending 4 a''} sequence. The camera performs a short trajectory from (a) to (b). In (c) we represent both frames and intermediate camera poses. (d) Number of frames seeing each surface point, with GT unobserved areas shown in gray. (e) We managed to reconstruct a curved section of the colon. (f) Our method plausibly estimates the wall of the colon at the right of camera (b), although it was never seen in the images.}
    \label{fig:t4v1}
\end{figure} 
We provide a qualitative result in \fig{with_without} and additional ones in the supplementary material. Note that the watertight prior inherent to an SDF allows the network to hallucinate unseen areas. Remarkably, these unsurveyed areas continue the tubular shape of the colon and we found them to be mostly accurate when compared to the ground truth.  For example, the curved areas of the colon where a wall is occluded behind the corner of the curve is reconstructed, as shown in \fig{t4v1}. This ability to ``fill in'' observation gaps may be useful in providing the endoscopist with an estimate of the percentage of unsurveyed area during a procedure.

We hypothesize that this desirable behavior stems from the fact that the network learns an empirical shape prior from the observed anatomy of the colon. However, we don't expect this behavior to hold for distant unseen parts, but only for regions closer than 20~mm to one observation. \changed{In the last rows} of \tab{metrics}, we compute accuracy metrics for this \textit{extended} region. It includes not only surveyed areas, but also neighboring areas that were not observed.
 
\section{Conclusion}

We have presented a method for 3D dense multi-view reconstruction from endoscopic images. We are the first to show that neural radiance fields can be used to obtain accurate dense reconstructions of colon sections of significant length. At the heart of our approach, is exploiting the correlation between depth and brightness. We have observed that, without it, neural reconstruction fails. 

\changed{The current method could be used offline for post-exploration coverage analysis and endoscopist training. But real-time performance could be achieved in the future as
the new NeuS2 \cite{wang2022neus2} converges in minutes, enabling automatic coverage reporting.} Similar to other reconstruction methods, for now our approach works in areas of the colon where there is little deformation. Several sub-maps of non-deformed areas can be created if necessary. However, this limitation could be overcome by adopting the deformable NeRFs formalism~\cite{park2021nerfies}.

\subsection*{Acknowledgement}

This work was supported by EU-H2020 grant 863146: ENDOMAPPER, Spanish government grants PID2021-127685NB-I00 and FPU20/06782 and by Aragón government grant DGA\_T45-17R.

\bibliographystyle{splncs04}
\bibliography{bib/main}

\clearpage
\appendix

\begin{center}
    {\Large\bfseries Supplementary material}
\end{center}

\begin{figure}
    \centering
    \begin{tabular}{ccc}
        (a) First frame & (c) Ground Truth & (e) 3D Reconstruction \\
        \includegraphics[width=0.25\textwidth]{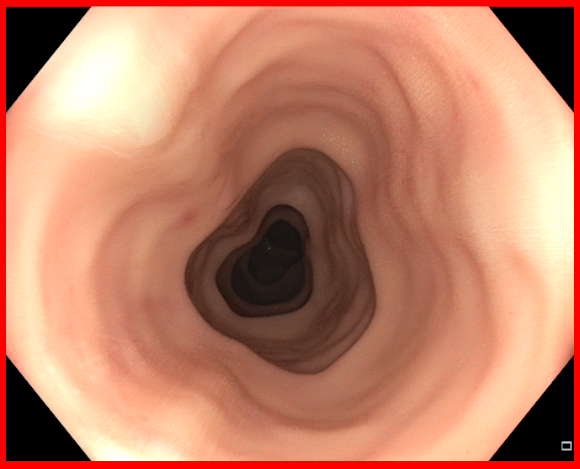} & 
        \includegraphics[width=0.25\textwidth]{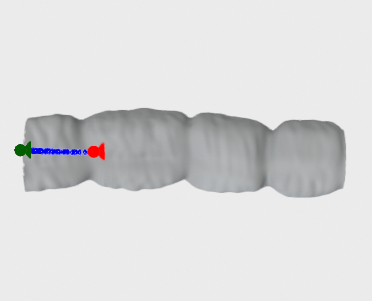} &
        \includegraphics[width=0.25\textwidth]{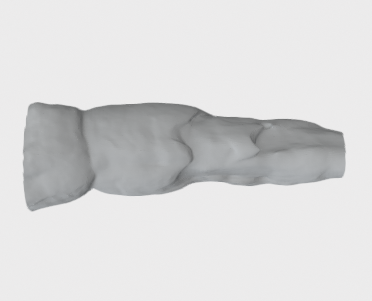} \\
        (b) Last frame & (d) Surveyed (GT) & (f) Error \\
        \includegraphics[width=0.25\textwidth]{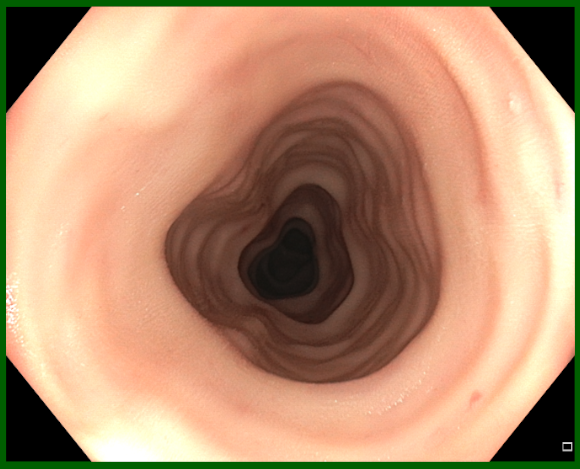} &
        \includegraphics[width=0.25\textwidth]{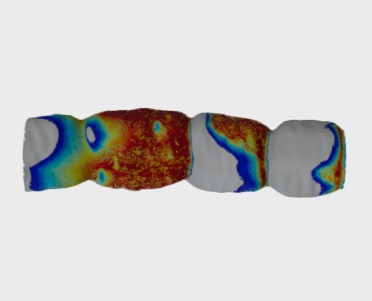} &
        \includegraphics[width=0.25\textwidth]{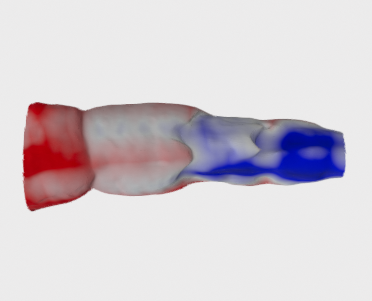} \\
        & \includegraphics[width=0.25\textwidth]{fig/colorbar_jet.jpg} & 
        \includegraphics[width=0.25\textwidth]{fig/colorbar_seismic.jpg} \\
        & 1 \quad \qquad \;\; \qquad \quad 20 & -1~cm \, \quad 0 \, \quad +1~cm
        
    \end{tabular}\fcaption{Reconstructing with low parallax}{Results on \emph{``Transcending 1 a''} sequence. (c) The camera travels in a straight line, covering less than a third of the section. As shown in (a) and (b), the haustra completely hide the background walls. (e) Consequently, the reconstruction underestimates the diameter of the end of the tube. However, the three characteristic folds in our reconstructed colon match the ground truth in number and location. In addition, areas observed multiple times ---red in (d)--- are reconstructed with high accuracy ---gray in (f).}
    \label{fig:t1v1}
\end{figure} 

\begin{figure}[!b]
    \centering
    \begin{tabular}{cccccc}
        Input \& GT & N = 2500 & N = 5000 & N = 12500 & N = 15000 & N = 65000 \\
        \includegraphics[width=0.16\textwidth]{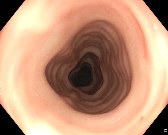} &
        \includegraphics[width=0.16\textwidth]{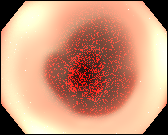} &
        \includegraphics[width=0.16\textwidth]{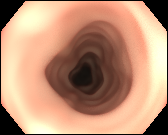} &
        \includegraphics[width=0.16\textwidth]{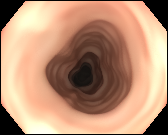} &
        \includegraphics[width=0.16\textwidth]{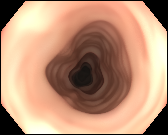} &
        \includegraphics[width=0.16\textwidth]{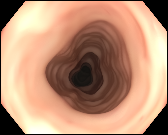}
        \\
        \includegraphics[width=0.16\textwidth]{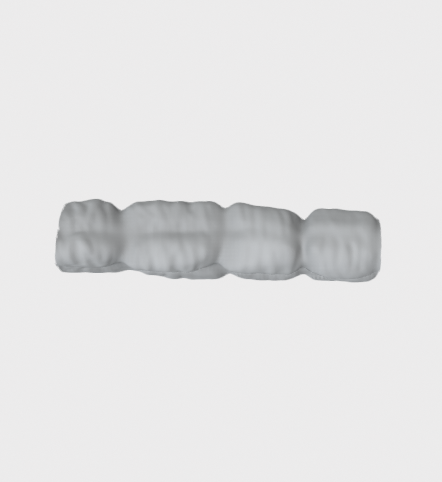} &
        \includegraphics[width=0.16\textwidth]{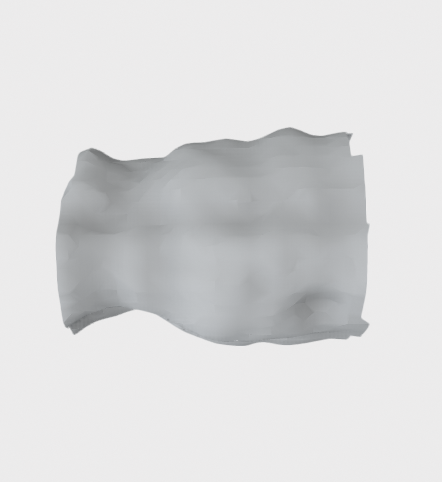} &
        \includegraphics[width=0.16\textwidth]{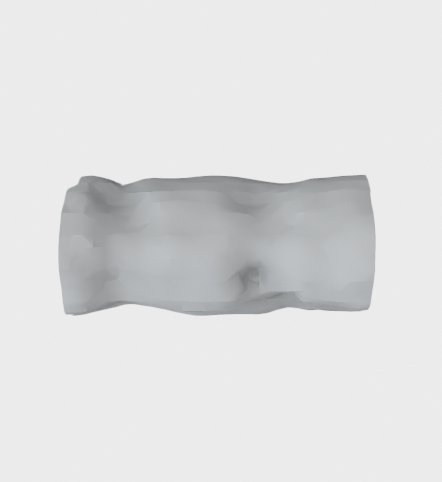} &
        \includegraphics[width=0.16\textwidth]{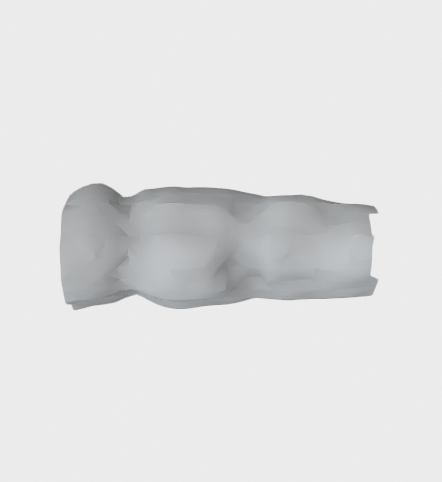} &
        \includegraphics[width=0.16\textwidth]{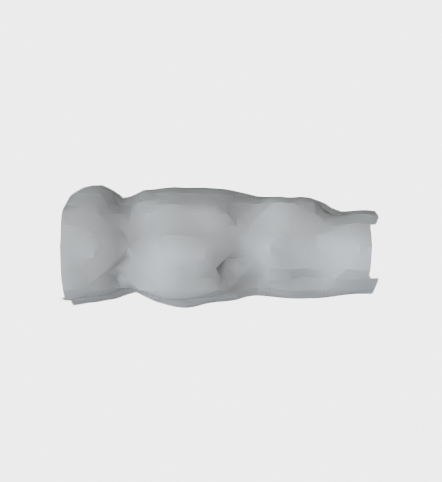} &
        \includegraphics[width=0.16\textwidth]{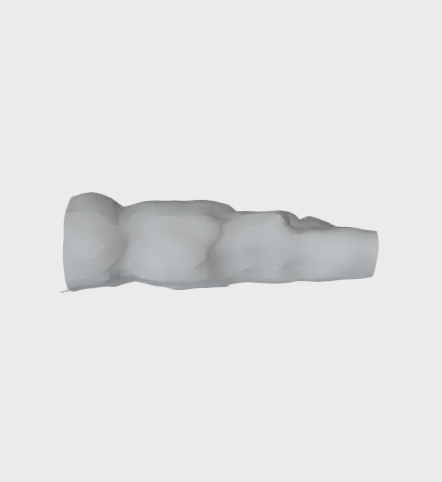}
        
    \end{tabular}\fcaption{Reconstruction convergence}{Results on \emph{``Transcending 1 a''} sequence. We show the intermediate results for N optimisation iterations. We see how the reconstruction converges quickly. In 65k iterations we already have a reasonable solution, compared to the 300k iterations proposed by the authors of NeuS.}
    \label{fig:convergence}
\end{figure} 

\begin{figure}
    \centering
    \begin{tabular}{ccc}
        (a) First frame & (c) Ground Truth & (e) 3D Reconstruction \\
        \includegraphics[width=0.25\textwidth]{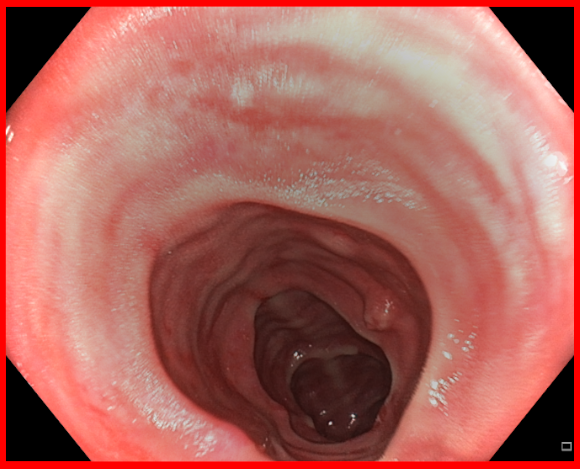} & 
        \includegraphics[width=0.25\textwidth]{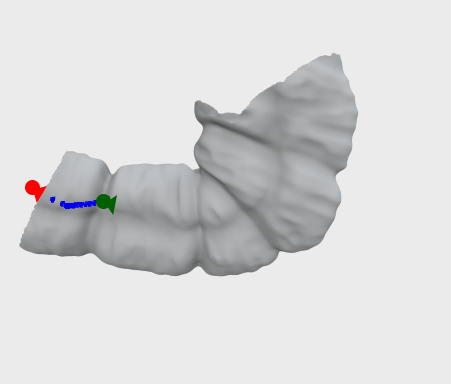} &
        \includegraphics[width=0.25\textwidth]{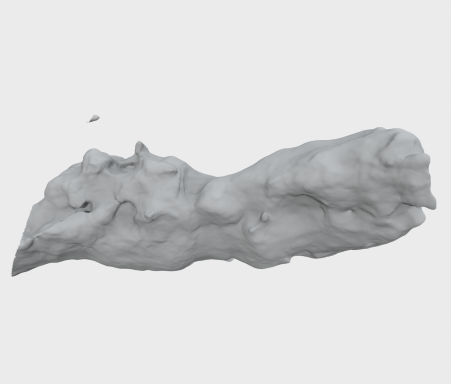} \\
        (b) Last frame & (d) Surveyed (GT) & (f) Error \\
        \includegraphics[width=0.25\textwidth]{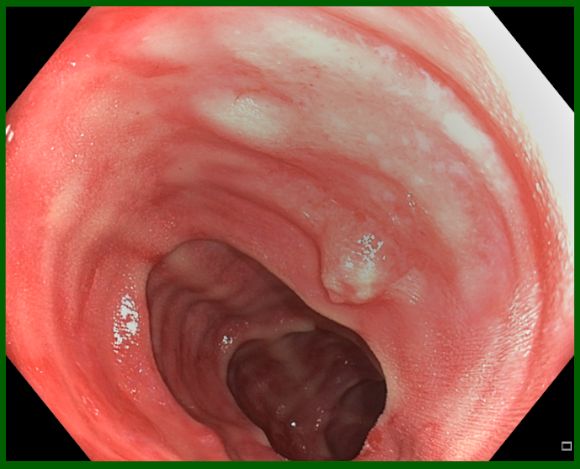} &
        \includegraphics[width=0.25\textwidth]{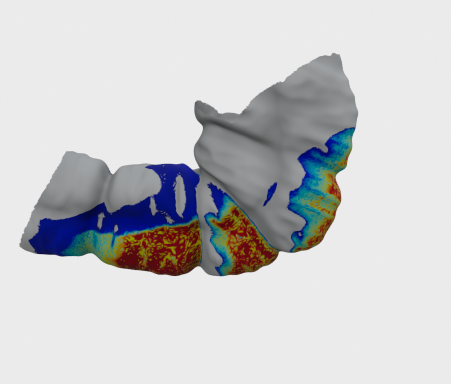} &
        \includegraphics[width=0.25\textwidth]{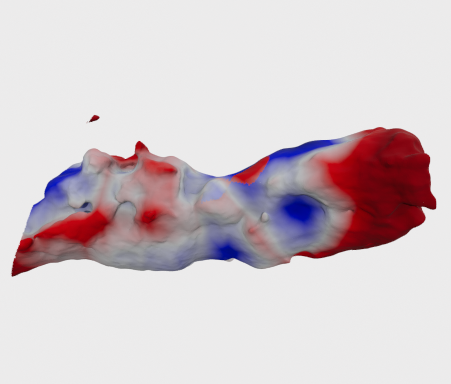} \\
        & \includegraphics[width=0.25\textwidth]{fig/colorbar_jet.jpg} & 
        \includegraphics[width=0.25\textwidth]{fig/colorbar_seismic.jpg} \\
        & 1 \quad \qquad \;\; \qquad \quad 20 & -1~cm \, \quad 0 \, \quad +1~cm
        
    \end{tabular}\fcaption{Reconstructing congruent shapes}{Results on \emph{``Descending 4 a''} sequence. (c) The camera takes the most challenging route: the shortest translation of the sequences tested, turning towards the right wall. (d) This results in very poor coverage, especially to the left of the camera. The curve to the left is never seen. (e) In this way we check that our method only  ``hallucinates'' partially observed areas, based on the structure of the region it has actually observed. As a result, the reconstruction continues as a straight tube. Again, areas observed multiple times ---red in (d)--- are reconstructed with high accuracy ---gray in (f).}
    \label{fig:d4v2}
\end{figure} \begin{figure}
    \centering
    \begin{tabular}{ccc}
        \includegraphics[width=0.25\textwidth]{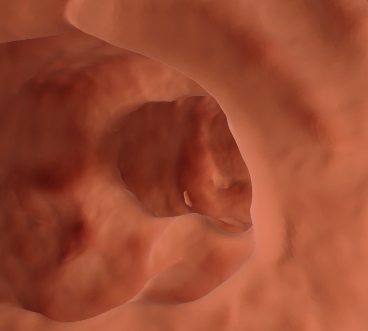} & 
        \includegraphics[width=0.25\textwidth]{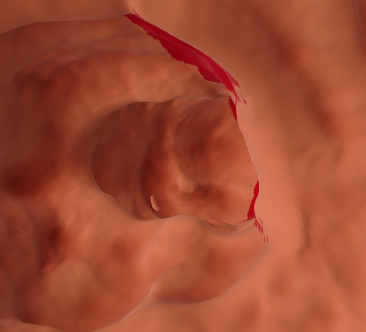} &
        \includegraphics[width=0.25\textwidth]{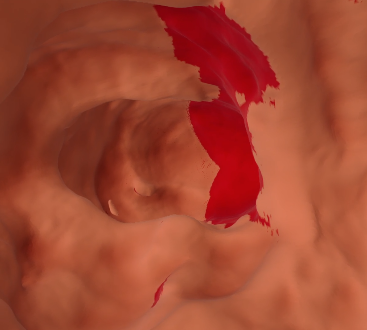} \\
        (a) & (b) & (c)
    \end{tabular}\fcaption{Post-intervention 3D visualization}{Results on \emph{``Transcending 4 a''} sequence. Our reconstructions would allow physicians to revisit the area explored during the endoscopy. This opens the door to augmented reality (AR) in post-intervention diagnostics. As an example, we show a visualisation of the area not surveyed during the procedure, marked in red. After inspecting a colon section, our watertight surface displays (a) visualized and (b), (c) non-visualized mucosa. The doctor can analyse non-visualized areas and make decisions about subsequent exploration. A video of this demonstration is included in the supplementary material.}
    \label{fig:inside}
\end{figure}  

\end{document}